\documentclass[conference]{IEEEtran}
\IEEEoverridecommandlockouts
\usepackage{cite}
\usepackage{amsmath,amssymb,amsfonts}
\usepackage{algorithmic}
\usepackage{graphicx}
\usepackage{textcomp}
\usepackage{xcolor}

\usepackage[colorlinks=true,
linkcolor=black,
anchorcolor=black,
citecolor=black, 
urlcolor=blue
]{hyperref}

\newcommand{\DOI}[1]{doi: \href{https://doi.org/#1}{#1}}

\def\BibTeX{{\rm B\kern-.05em{\sc i\kern-.025em b}\kern-.08em
    T\kern-.1667em\lower.7ex\hbox{E}\kern-.125emX}}
\begin{document}

\title{Hierarchical Salient Patch Identification for Interpretable Fundus Disease Localization\\
\thanks{* Corresponding authors}
}

\author{
	\textbf{Yitao Peng\textsuperscript{2}},
	\textbf{Lianghua He\textsuperscript{1,2*}},
	\textbf{Die Hu\textsuperscript{3*}}\\
	\textsuperscript{1}Shanghai Eye Disease Prevention and Treatment Center, Shanghai 200040, China\\
	\textsuperscript{2}School of Electronic and Information Engineering, Tongji University, Shanghai 201804, China\\
	\textsuperscript{3}School of Information Science and Technology, Fudan University, Shanghai 200433, China\\
	Email: pyt@tongji.edu.cn, helianghua@tongji.edu.cn, hudie@fudan.edu.cn
}

\maketitle

\begin{abstract}
With the widespread application of deep learning technology in medical image analysis, the effective explanation of model predictions and improvement of diagnostic accuracy have become urgent problems that need to be solved. Attribution methods have become key tools to help doctors better understand the diagnostic basis of models, and are used to explain and localize diseases in medical images. However, previous methods suffer from inaccurate and incomplete localization problems for fundus diseases with complex and diverse structures. To solve these problems, we propose a weakly supervised interpretable fundus disease localization method called hierarchical salient patch identification (HSPI) that can achieve interpretable disease localization using only image-level labels and a neural network classifier (NNC). First, we propose salient patch identification (SPI), which divides the image into several patches and optimizes consistency loss to identify which patch in the input image is most important for the network's prediction, in order to locate the disease. Second, we propose a hierarchical identification strategy to force SPI to analyze the importance of different areas to neural network classifier's prediction to comprehensively locate disease areas. Conditional peak focusing is then introduced to ensure that the mask vector can accurately locate the disease area. Finally, we propose patch selection based on multi-sized intersections to filter out incorrectly or additionally identified non-disease regions. We conduct disease localization experiments on fundus image datasets and achieve the best performance on multiple evaluation metrics compared to previous interpretable attribution methods. Additional ablation studies are conducted to verify the effectiveness of each method.
\end{abstract}


\section{Introduction} \label{Section_1}

The diagnosis of fundus disease \cite{fu2019palm,gulshan2016development} is critical because it is a leading cause of vision loss and blindness worldwide. However, the imaging features of fundus diseases are complex and variable \cite{cen2021automatic}, and the manifestations of different diseases vary greatly, posing significant challenges for the automated identification of fundus diseases. Traditional fundus image segmentation technology relies on accurate pixel-level labels \cite{pan2022label} to train models to identify diseased areas. However, obtaining accurate annotations of fundus disease images is time-consuming and expensive \cite{goutam2022comprehensive}. The existing weakly supervised learning methods require only image-level labels for disease localization \cite{chen2022c}. Although these methods reduce reliance on precise annotations, they can usually only locate a part of the disease area rather than comprehensively cover the entire disease area. This incomplete localization may lead to a missed diagnoses and thereby affect treatment. Furthermore, the existing weakly supervised localization methods face challenges in terms of interpretability \cite{wargnier2023weakly}. Their black-box nature makes it difficult for doctors to trust the model predictions, thus limiting their clinical application \cite{wysocki2023assessing}. Therefore, developing a weakly supervised learning method that can improve both the completeness and interpretability of disease localization in fundus images is an urgent problem that needs to be solved.

\begin{figure}[!t]
	\centerline{\includegraphics[width=\columnwidth]{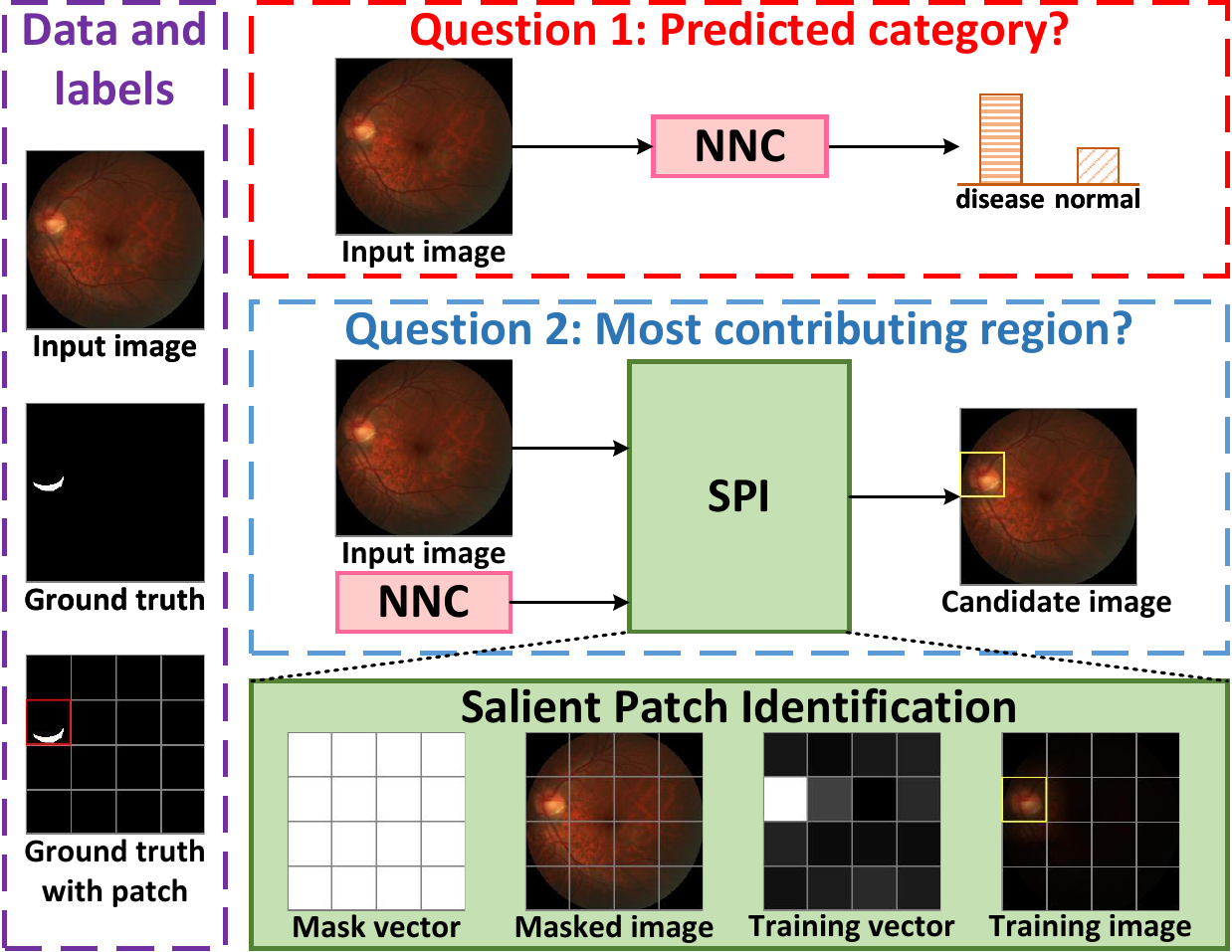}}
	\caption{The principle of weakly supervised interpretable fundus disease localization using SPI. Question 1: Which category does the NNC predict for the input image? Question 2: Which region contributes most to the NNC prediction? NNC and SPI can respectively answer Question 1 and Question 2. When the NNC predicts that the input image is diseased, the region that contributes most to the NNC prediction is identified as the diseased area. SPI uses a mask vector to divide the input image into multiple patches to learn the patch that contributes most to the NNC prediction.}
	\label{fig_1}
\end{figure}

Attribution methods \cite{selvaraju2017grad,chattopadhay2018grad} provide neural network classifiers (NNCs) with the ability to perform explainable localization using only image-level labels. Because NNCs usually use only one area of the image to make predictions, traditional attribution methods \cite{wang2020score,fu2020axiom} often locate only a part of the disease area and cannot cover the entire disease area. However, the foci of fundus diseases are often distributed across multiple locations, which can lead to incomplete localization. Because the NNC may use both diseased and surrounding non-diseased areas when making disease predictions, the results generated by the attribution method may locate normal areas and thereby lead to less accurate positioning.

To address these issues, we first propose SPI. Figure \ref{fig_1} illustrates this principle. The NNC is trained using image-level labels to predict fundus diseases. When the NNC predicts the input image as diseased, the area contributing the most to the NNC prediction is usually where the disease occurs. The SPI identifies the most important areas for NNC prediction to determine disease regions (yellow patch in Figure \ref{fig_1}), thereby achieving weakly supervised localization of fundus diseases. The SPI masks the input image through the upsampled mask vector to obtain the masked image, that is, the initial training image. SPI constrains the training and input images to be consistent at the output of the NNC to train the mask vector. The training vector retains the areas the NNC considers to contribute greatly to the prediction (white patch in the training vector) and removes the areas that contribute little to the prediction (black patches in the training vector), thus providing interpretability for the SPI recognition process \cite{peng2023hierarchical}.

Second, inspired by the process of doctors diagnosing diseases, we propose a hierarchical approach using the SPI, which leads to the development of the HSPI. After determining the most likely disease area in a medical image, doctors continue to examine other areas. Similarly, by masking the identified areas in the input image and the mask vector, the mask vector is trained again to identify new patches (i.e., disease areas) that make the greatest contribution to the NNC prediction, thereby comprehensively identifying the disease areas.

Third, to improve the accuracy of diseased area identification using mask vectors in SPI, we propose a learning convergence condition called conditional peak focusing (CPF). By ensuring that the training results of the mask vector comply with the CPF, the mask vector accurately retains the areas most important for prediction while eliminating irrelevant areas.

Finally, because different images may contain diseased regions of varying numbers and sizes, a fixed number of hierarchical identifications in the HSPI may result in too many or too few patches. Therefore, we propose patch selection based on multi-sized intersections (PSMI). This method filters the patches identified by the HSPI using mask vectors of different sizes, ensuring that only diseased areas are retained, while removing misidentified or redundant non-diseased areas.

Our key contributions are as follows:

\begin{itemize}
	\item We propose a weakly supervised interpretable fundus disease localization method called HSPI. This method requires image-level labels and an NNC. By adopting mask learning and hierarchical identification technologies, HSPI can accurately and comprehensively locate areas of fundus disease and provide interpretability.
	\item To enable the SPI to locate disease areas more accurately, we introduce CPF as the mask vector selection condition for training to ensure that the training vector ultimately focuses on the disease area most relevant for prediction.
	\item We propose PSMI to analyze whether each image patch contains a disease by combining mask vectors of different sizes. This method filters non-disease patches that are incorrectly or additionally identified, thereby improving the identification accuracy of disease patches.
\end{itemize}

\section{Related Work} \label{Section_2}
\subsection{Attribution Methods} \label{Section_2_1}
In medical image analysis, the interpretability of deep learning models is crucial. Attribution methods, including gradient-based \cite{sundararajan2017axiomatic}, perturbation-based \cite{petsiuk2018rise}, and class activation map (CAM-based) \cite{wang2020score}, can reveal how models understand complex medical images. Gradient-based methods reveal the contribution of features to predictions by calculating the gradient. The perturbation-based methods analyze predictions by perturbing the input data and observing changes in the outputs. CAM-based methods reveal how a network operates by visualizing class activation maps to demonstrate the attention.

However, these methods struggle with inaccurate and incomplete positioning in complex fundus images.

\begin{figure*}[!t]
	\centering
	{\includegraphics[width=1.0\linewidth]{{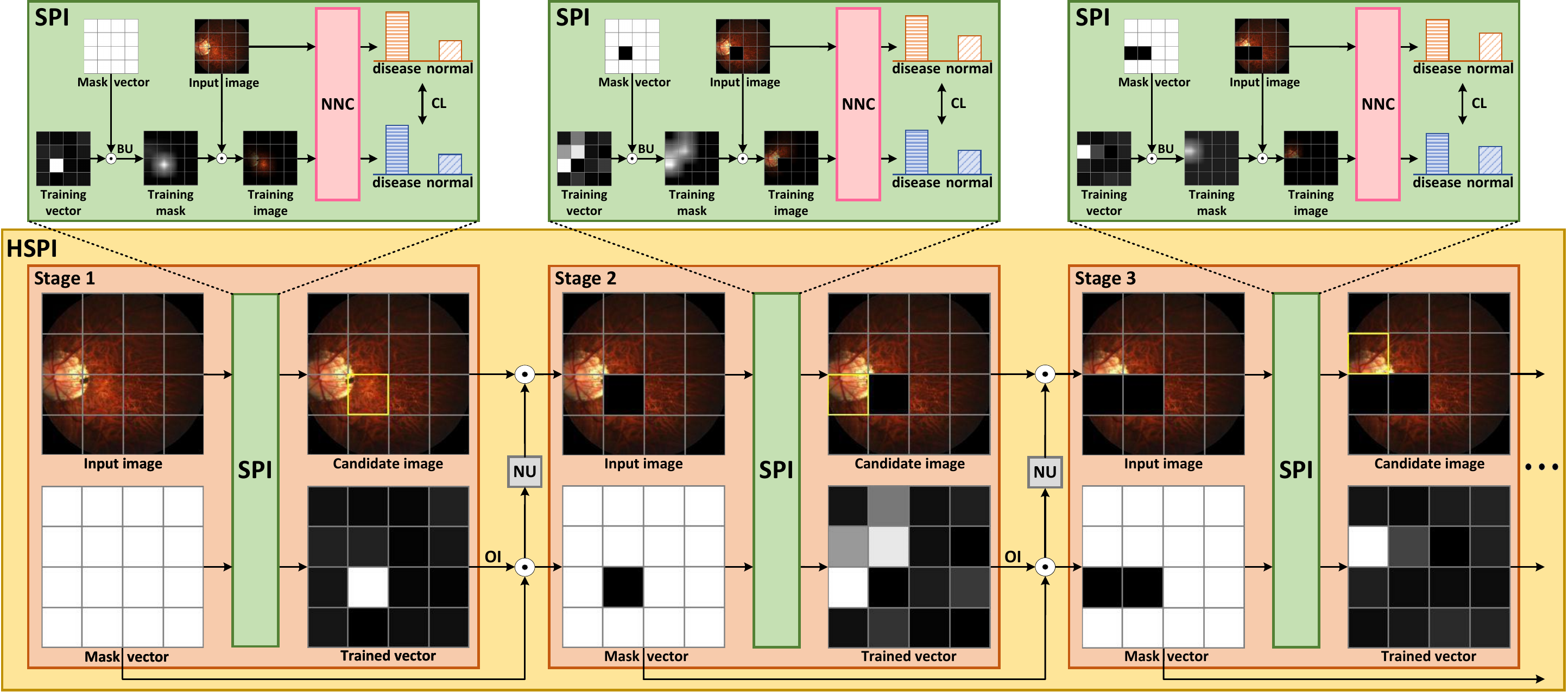}}%
	}
	\caption{The overall architecture of HSPI. In Stage 1, the original input image and mask vector are fed into SPI and learned by computing the consistency loss (CL) to produce candidate image and trained vector for NNC prediction. Then, process the mask vector, trained vector, and candidate image from Stage 1 to obtain the input image and mask vector for Stage 2. In Stages 2, 3, and so on, repeat the above process. HSPI finally identifies several candidate patches (yellow patches) for NNC prediction.}
	\label{fig_2}
\end{figure*}

\subsection{Weakly Supervised Semantic Segmentation} \label{Section_2_2}

In weakly supervised semantic segmentation (WSSS) \cite{chen2022c}, four types of weak labels have been explored: image-level labels \cite{chen2022c}, point annotations \cite{kim2023devil}, graffiti labels \cite{lin2016scribblesup}, and bounding boxes \cite{li2023sim}. Image-level labels are commonly used in medical image analysis. Hence, this study relied only on image-level supervision. Most current image-level supervised WSSS research is based on CAM, which locates key discriminant regions in conjunction with NNC \cite{chen2022c}. Because these WSSS methods rely on black-box models for segmentation, it is difficult to explain the reasoning process, making it hard for doctors to understand how the model localizes lesions.

To tackle these issues, this study proposes a positioning technology called HSPI, which aims to provide interpretable, accurate, and comprehensive positioning for fundus diseases to assist in the diagnosis of fundus imaging diseases.

\section{Methodology} \label{Section_3}

\subsection{Hierarchical Salient Patch Identification} \label{S3_1}

The overall structure of HSPI is shown in Figure \ref{fig_2}. We first introduce the process of SPI. Define: input image $X_{0} \in \mathbb{R}^{H \times W \times 3}$, all-one mask vector $E_{(i,0)} = [1]_{H_{i} \times W_{i}}$, mask vector $M_{(i,j)} \in \mathbb{R}^{H_{i} \times W_{i} \times 1}$, its matrix representation is as:
\begin{equation} \label{eq_1}
\begin{aligned}
M_{(i,j)} = [m^{(i,j)}_{(u,v)}]_{H_{i} \times  W_{i}}
\end{aligned}
\end{equation}
where $m^{(i,j)}_{(u,v)} \in [0,1]$, $m^{(i,j)}_{(u,v)}$ is a learnable variable initialized to $\zeta$; $m^{(i,j)}_{(u,v)}=0$ means complete shielding (black), and $m^{(i,j)}_{(u,v)}=1$ means complete retention (white); $i$ ($i \in \{1,2,...,I\}$) indicates that the mask vector belongs to the $i$-th size category (height $H_{i}$ and width $W_{i}$); $j$ ($j \in \{1,2,...,J\}$) represents the hierarchical identification currently in the $j$-th stage; $(u,v)$ represents the position of the element $m^{(i,j)}_{(u,v)}$ in the matrix $[m^{(i,j)}_{(u,v)}]_{H_{i} \times  W_{i}}$; $I$ and $J$ represent the number of dimension categories and hierarchical recognitions, respectively. Let $G_{B}(\cdot)$ and $G_{N}(\cdot)$ represent bilinear interpolation and nearest neighbor interpolation upsampling, respectively, which upsample the mask vector to the size of the input image ($H \times W$). $\Upsilon_{i} = \{(u,v)|u \in \{1, 2, ..., H_{i}\}, v \in \{1, 2, ..., W_{i}\} \}$. Let $T(M_{(i,j)})$ and $T(m^{(i,j)}_{(u,v)})$ represent the mask vector and its values after training, respectively. Then, the subscript for the position of the maximum value in $T(M_{(i,j)})$ is as follows:
\begin{equation} \label{eq_2}
\begin{aligned}
(u^{(i,j)}_{m}, v^{(i,j)}_{m}) = { \underset {(u,v) \in \Upsilon_{i}} { \operatorname {arg\,max} } \, T(m^{(i,j)}_{(u,v)})} 
\end{aligned}
\end{equation}

Let $\Lambda_{(i,j)} = \{(u^{(i,j)}_{m}, v^{(i,j)}_{m})\}$. The rigid local mask vector $E_{(i,j)}$ is generated according to $(u^{(i,j)}_{m}, v^{(i,j)}_{m})$ as:
\begin{equation} \label{eq_3}
\begin{aligned}
E_{(i,j)} = & \left\{[e^{(i,j)}_{(u,v)}]_{H_{i} \times W_{i}} \middle| e^{(i,j)}_{(u,v)}=\begin{cases} 0,  (u, v) \in \Lambda_{(i,j)} \\ 1,  (u, v) \notin \Lambda_{(i,j)}\end{cases}\right\}
\end{aligned}
\end{equation}

In stage $j$, according to the rigid local mask vector $E_{(i,j)}$, the most critical patch for prediction identified by $E_{(i,j)}$ is located. The patch with a value of 0 in $G_{N}(E_{(i,j)})$ represents the diseased area in the current input image obtained by SPI in stage $j$ that is most conducive for NNC prediction.

In SPI, we introduced the two losses to train $M_{(i,j)}$ to become $T(M_{(i,j)})$. For an input image $X \in \mathbb{R}^{H \times W \times 3}$ and a mask $M \in \mathbb{R}^{H \times W \times 1}$, the similarity loss $L_{s}$ represents the square of the difference in logits produced by inputting the image $X$ and the masked image $MX$ into the NNC $f(\cdot)$ as:
\begin{equation} \label{eq_4}
\begin{aligned}
L_{s}(X, M) = ||f(X) - f(MX) ||_{2}^{2}
\end{aligned}
\end{equation}
where $||\cdot||_{2}$ is L2-norm. For $M_{(i,j)}$, the mask loss $L_{m}$ is:
\begin{equation} \label{eq_5}
\begin{aligned}
L_{m}(M_{(i,j)}) = \frac{1}{{|H_{i} W_{i}|}}\sum^{H_{i}}_{u=1}\sum^{W_{i}}_{v=1} |m^{(i,j)}_{(u,v)}|
\end{aligned}
\end{equation}

The rigid global mask vector $\Omega_{(i,j)}$ of stage $j$ is generated based on all the rigid local mask vectors $\{E_{(i,k)}\}^{j}_{k=0}$ produced before stage $j$ as follows:
\begin{equation} \label{eq_6}
\begin{aligned}
\Omega_{(i,j)} = \prod_{k=0}^{j} E_{(i,k)}
\end{aligned}
\end{equation}
where $\Omega_{(i,j)}$ describes the $j$ patches (0 regions) identified by SPI in the input image at stage $j$ and before, which contribute to NNC prediction. In this study, we proposed a hierarchical identification technology. In stage $j$ of HSPI, the $j-1$ areas that have been previously identified are masked to ensure that subsequent SPI can identify the $j$-th prediction-critical patch located at a different location. The input image $X_{0}$ is masked using the upsampled rigid global mask vector $G_{N}(\Omega_{(i,j-1)})$ to generate the input image $X_{(i,j)}$ for stage $j$, as follows:
\begin{equation} \label{eq_7}
\begin{aligned}
X_{(i,j)} = G_{N}(\Omega_{(i,j-1)})X_{0}
\end{aligned}
\end{equation}

The mask vector $M_{(i,j)}$ is then trained to identify what NNC $f(\cdot)$ considers to be the $j$-th most important patch in the input image $X_{0}$ by consistency loss (CL) $L_{c}$ in SPI. This formula is defined as follows:
\begin{equation} \label{eq_8}
\begin{aligned}
L_{c}(X_{(i,j)}, M_{(i,j)}) & = L_{s}(X_{(i,j)}, G_{B}(\Omega_{(i, j-1)}M_{(i,j)})) \\
& + \lambda_{(i,j)} L_{m}(M_{(i,j)})
\end{aligned}
\end{equation}
where $\lambda_{(i,j)}$ is a balancing factor, $M_{(i,j)}$ is trainable, and $X_{(i,j)}$ and $\Omega_{(i, j-1)}$ are both fixed. Note that when training (\ref{eq_8}), bilinear interpolation upsampling $G_{B}(\cdot)$ is used instead of nearest neighbor interpolation upsampling $G_{N}(\cdot)$ to generate the training mask. The purpose is to avoid introducing sharp edges, thus making the training of the mask vector stable \cite{petsiuk2018rise}.

The steps for HSPI are shown in Figure \ref{fig_2}. In Stage 1, the input image $X_{(i,1)}$ ($X_{(i,1)}=G_{N}(\Omega_{(i,0)})X_{0} = X_{0}$) and the rigid global mask vector $\Omega_{(i,0)}$ ($\Omega_{(i,0)} = E_{(i,0)}$) are input into SPI. In Stage 1 of SPI, the training vector $M_{(i,1)}$ (trainable) and $\Omega_{(i,0)}$ (fixed) are multiplied elementwise. The results are then applied using bilinear interpolation upsampling (BU) to generate a training mask $G_{B}(\Omega_{(i,0)}M_{(i,1)})$ of the same size as the input image $X_{(i,1)}$. Then, $X_{(i,1)}$ is element-wise multiplied with the training mask $G_{B}(\Omega_{(i,0)}M_{(i,1)})$ to obtain the training image $G_{B}(\Omega_{(i,0)}M_{(i,1)})X_{(i,1)}$. Next, the input image $X_{(i,1)}$ and the training image $G_{B}(\Omega_{(i,0)}M_{(i,1)})X_{(i,1)}$ are input into NNC, and then the consistency loss (CL) is optimized to train the training vector $M_{(i,1)}$. Subsequently, both the final trained vector $T(M_{(i,1)})$ and the candidate image are obtained from Stage 1. The most salient patch in $T(M_{(i,1)})$ constitute the candidate patch (yellow patch) in the candidate image.

Subsequently, one-hot inversion (OI) is performed on $T(M_{(i,1)})$ in Stage 1 to obtain the rigid local mask vector $E_{(i,1)}$. The OI operation involves two steps: one-hot encoding and binary inversion. Elementwise multiplication of $E_{(i, 1)}$ and $\Omega_{(i,0)}$ yields the rigid global mask vector $\Omega_{(i,1)}=E_{(i, 1)}\Omega_{(i,0)}$ for Stage 2. $\Omega_{(i,1)}$ is then upsampled using nearest neighbor interpolation (NU) to produce mask $G_{N}(\Omega_{(i,1)})$, which is elementwise multiplied with the candidate image to produce the masked image $G_{N}(\Omega_{(i,1)})X_{0}$, that is, the input image $X_{(i,2)} = G_{N}(\Omega_{(i,1)})X_{0}$ of Stage 2. In Stages 2, 3, ..., and $J$, the operation of Stage 1 is repeated, and $J$ candidate images are obtained. Each candidate image indicates a patch (yellow patch) that represents the area in the corresponding stage input image that has the highest contribution to the prediction. HSPI integrates these patches to obtain the positioning result.

\subsection{Conditional Peak Focusing} \label{S3_2}

\begin{figure}[!t]
	\centering
	{\includegraphics[width=1.0\linewidth]{{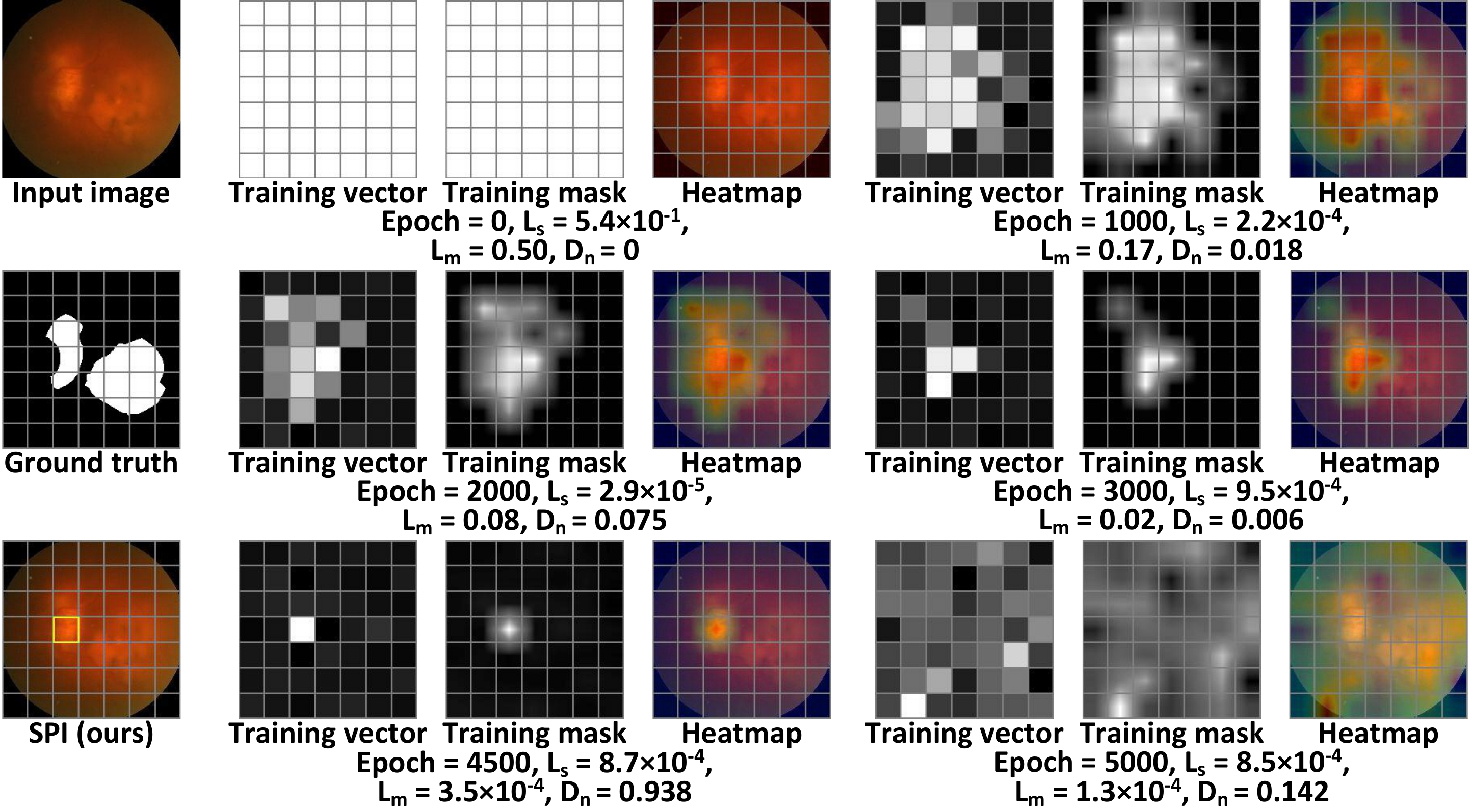}}%
	}
	\caption{An example of training mask vector in SPI to identify disease area for prediction. The result of SPI training mask vector, as well as the corresponding training period (Epoch), similarity loss $L_{s}$, mask loss $L_{m}$, and maximum-second difference $D_{n}$.}
	\label{cpf}
\end{figure}

In the previous section, we introduced the SPI method for training the mask vector $M_{(i,j)}$ to obtain the trained vector $T(M_{(i,j)})$ by optimizing consistency loss $L_{c}$. Because SPI must continuously optimize $L_{c}$, the training period (Epoch) during which the mask vector is trained is particularly important. Let the current training period is $n$, the total training period is $n^{t}$, and the learning rate is $\eta$. Figure \ref{cpf} shows the process of SPI optimization $L_{c}$ as an example. If the training period is small (Epoch $n$ = 0 to 3000), the similarity loss $L_{s}$ in $L_{c}$ is large. This results in inconsistency between the key prediction contribution information of the training and original input images. The training vector may not have reached a convergence state, resulting in an inaccurately identified key area. In contrast, if the training period is too long (Epoch $n$ = 5000), because the mask loss $L_{m}$ in $L_{c}$ gradually decreases, all position values of the mask vector continue to be reduced, and the peak value may even be smoothed. Therefore, it is crucial to determine the most appropriate training period to ensure that the training vector can effectively highlight the peak, and more accurately and intuitively identify the areas with the highest contribution to the prediction of the NNC. Therefore, we propose a method known as conditional peak focusing (CPF). This method determines the most appropriate training period by setting constraints for the training of mask vectors, ensuring that the trained vector obtained is ideal when the following three conditions are met:

Firstly, for the training mask vector $M_{(i,j)}$, in order to ensure that the training image $G_{B}(\Omega_{(i,j-1)}M_{(i,j)})X_{(i,j)}$ has the same most critical prediction contribution area as the input image $X_{(i,j)}$, the similarity loss $L_{s}$ needs to be small. Therefore, $L_{s}$ should be maintained below the preset small threshold $\alpha_{(i,j)}$ as follows:
\begin{equation} \label{eq_9}
\begin{aligned}
L_{s}(X_{(i,j)}, G_{B}(\Omega_{(i,j-1)}M_{(i,j)})) < \alpha_{(i,j)}
\end{aligned}
\end{equation}

Secondly, for the training mask vector $M_{(i,j)}$ to remove areas irrelevant to the prediction of the NNC, the sum of mask values must be small. That is, mask loss $L_{m}$ should be smaller than the preset small threshold $\beta_{(i,j)}$ as follows:
\begin{equation} \label{eq_10}
\begin{aligned}
L_{m}(M_{(i,j)}) < \beta_{(i,j)}
\end{aligned}
\end{equation}

Finally, for the trained vector $T(M_{(i,j)})$ to retain the peak area to reflect the prediction-critical area, we select the training vector $M_{(i,j)}$ to satisfy the maximum difference between the maximum and submaximum values. $M^{n}_{(i,j)}$ represents the result when the training period of $M_{(i,j)}$ reaches round $n$, and $MAX(M^{n}_{(i,j)})$ and $SECOND(M^{n}_{(i,j)})$ represent the maximum and second largest element values of $M^{n}_{(i,j)}$, respectively. Let maximum-second difference $D_{n}(i,j) = MAX(M^{n}_{(i,j)}) - SECOND(M^{n}_{(i,j)})$. $N_{e}$ is the set of all $n$ that satisfy (\ref{eq_9}) and (\ref{eq_10}). When the training period reaches $n^{m}$, $D_{n}(i, j)$ reaches the maximum value, $D_{n^{m}}(i, j)$, which satisfies:
\begin{equation} \label{eq_11}
\begin{aligned}
D_{n^{m}}(i, j) \geq D_{n}(i, j), \forall n \in N_{e}
\end{aligned}
\end{equation}

As shown in Figure \ref{cpf}, when $n=n^{m}=4500$; $L_{s}$, $L_{m}$, and $D_{n}$ comply with CPF; and the training vector reaches its best, the diseased areas have been successfully localized.

\subsection{Patch Selection Based on Multi-sized Intersections}  \label{S3_3}

HSPI hierarchically identifies $J$ patches in the input image that contribute to the prediction by setting $J$ rounds. However, the number of patches that contribute to the prediction may differ in different disease images, if a fixed $J$ value is artificially set, it may lead to an inappropriate number of identified patches. For an image with many disease areas, a $J$ value that is small will result in incompletely identified disease areas, whereas for an image with few disease areas, a $J$ value that is large will lead to the identification of additional non-disease areas. Therefore, to identify the appropriate number of patches in different images, we propose a method of patch selection based on multi-sized intersections (PSMI).

The PSMI technology is as follows: the mask vectors of different sizes ($\{H_{i} \times W_{i}\}_{i=1}^{I}$) are set to calculate the $J$ most critical patches identified by HSPI under different size divisions. Using the patches identified by the $H_{i_{b}} \times W_{i_{b}}$ trained vector $\{T(M_{(i_{b},j)})\}_{j=1}^{J}$ as a benchmark, patches identified by trained vectors $\{\{T(M_{(i,j_{o})})\}_{j_{o}=1}^{J}\}_{i \neq i_{b}}$ of other sizes ($\{H_{i} \times W_{i}\}_{i \neq i_{b}}$) are examined. The number of other patches located by $\{T(M_{(i_{b},j)})\}_{j=1}^{J}$ that fall into it each patch identified by $\{\{T(M_{(i,j_{o})})\}_{j_{o}=1}^{J}\}_{i \neq i_{b}}$ is calculated. If the number is greater than the threshold, it means that in the case of division of other sizes, the NNC determines that the area contains disease, and the patch is retained; otherwise, it is removed.

Let $\mathbb{E} = [1]_{H \times W}$ and $\mathbb{O} = [0]_{H \times W}$. $T(M_{(i,j)})$ is processed by OI to yield $E_{(i,j)}$. The patch identified by $T(M_{(i,j)})$ can be expressed as an area with a value of 1 in $q_{(i,j)}$ as:
\begin{equation} \label{eq_12}
\begin{aligned}
q_{(i,j)} = \mathbb{E} - G_{N}(E_{(i,j)})
\end{aligned}
\end{equation}

Use $\Phi((i_{1},j_{1}), (i_{2}, j_{2}))$ to indicate whether the prediction-critical patches identified as trained vectors $T(M_{(i_{1},j_{1})})$ and $T(M_{(i_{2},j_{2})})$ intersect. Its formula is as follows:
\begin{equation} \label{eq_13}
\begin{aligned}
\Phi((i_{1},j_{1}), (i_{2}, j_{2})) = MAX(q_{(i_{1},j_{1})}q_{(i_{2},j_{2})})
\end{aligned}
\end{equation}
where (\ref{eq_13}) equals 1, which indicates an intersection, and 0, which indicates no intersection. Subsequently, the number of intersections of patches generated by $T(M_{(i_{b}, j)})$ and $\{\{T(M_{(i, j_{o})})\}_{j_{o}=1}^{J}\}_{i \neq i_{b}}$ determines whether $q_{(i_{b},j)}$ is retained as:
\begin{equation} \label{eq_14}
\begin{aligned}
s_{(i_{b},j)}=\begin{cases} q_{(i_{b},j)},  (\sum_{i \neq i_{b}} \mathop{\max}\limits_{1 \leq j_{o} \leq J} \Phi((i_{b},j), (i, j_{o}))) \geq \epsilon \\ \mathbb{O}, \ others \end{cases}
\end{aligned}
\end{equation}
where $\epsilon$ is the threshold. Non-prediction-critical patch in stage $j$ are removed through the selection of the above threshold $\epsilon$. The final disease location result $S_{i_{b}}$ is obtained by filtering the output of HSPI using PSMI, as follows:
\begin{equation} \label{eq_15}
\begin{aligned}
S_{i_{b}} = \sum_{j=1}^{J}s_{(i_{b},j)}
\end{aligned}
\end{equation}

\section{Experiment}

\subsection{Datasets and Baselines}
We experimented on the PALM \cite{fu2019palm} and EyePACS \cite{gulshan2016development} datasets. PALM dataset contains 1,200 standardized fundus images divided into three groups: training, validation, and testing, each containing 400 images. Some images also contained segmentation labels for lesion regions. EyePACS dataset contains 35,126 diabetic retinal fundus images. They are divided into five levels from 0 to 4, covering five categories, and provide corresponding images and their category labels.

We used various classical or advanced attribution methods as baselines that could perform interpretable weakly supervised localization using only image-level labels and an NNC. We compared eight CAM-based attribution methods: Grad-CAM \cite{selvaraju2017grad}, Grad-CAM++ \cite{chattopadhay2018grad}, Score-CAM \cite{wang2020score}, XGrad-CAM \cite{fu2020axiom}, Eigen-CAM \cite{muhammad2020eigen}, Eigen Grad-CAM \cite{duguaesescu2022evaluation}, Ablation-CAM \cite{ramaswamy2020ablation}, and Layer-CAM \cite{jiang2021layercam}. We also compared seven other types of attribution methods: LIME \cite{ribeiro2016should}, G-LIME \cite{li2023g}, RISE \cite{petsiuk2018rise}, Gradient \cite{simonyan2013deep}, IG \cite{sundararajan2017axiomatic}, DeepLIFT \cite{shrikumar2017learning}, and SHAP \cite{lundberg2017unified}.

\begin{figure*}[!t]
	\centering
	{\includegraphics[width=1.0\linewidth]{{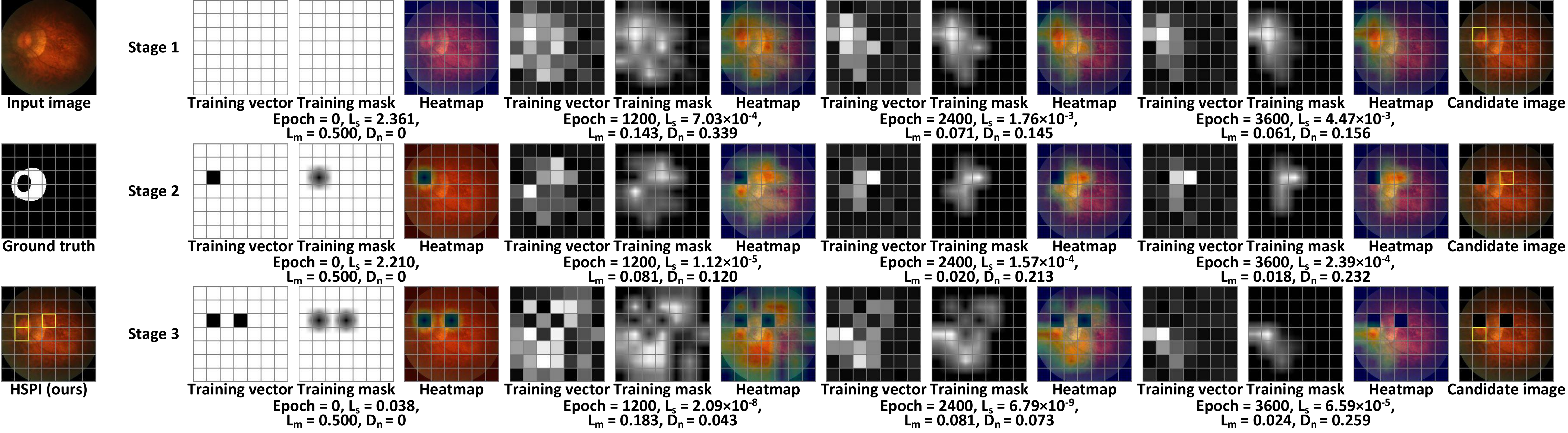}}%
	}
	\caption{An example of HSPI identifying disease areas (yellow patches) for prediction. It shows the results of HSPI training mask vectors using SPI at different stages, as well as the corresponding training period (Epoch), similarity loss $L_{s}$, mask loss $L_{m}$, and maximum-second difference $D_{n}$.}
	\label{process}
\end{figure*}

\subsection{Evaluation}

To evaluate the localization ability of the proposed method, we utilized five key metrics: F1-score (F1) \cite{cen2021automatic}, Positive Predictive Value (PPV) \cite{peng2023hierarchical}, Specificity (SP) \cite{cen2021automatic}, Average Surface Distance (ASD) \cite{chen2022c}, and Proportion \cite{peng2023hierarchical}. To calculate four of the metrics F1, PPV, SP, and ASD, we transformed the saliency maps generated by the baseline methods into binary pseudo-masks based on the threshold corresponding to the highest average F1 value. The formulas are as follows: $F1= \frac{2TP}{FP+2TP+FN}$, $PPV= \frac{TP}{TP+FP}$, and $SP= \frac{TN}{TN+FP}$, where TP, TN, FP, and FN represent true positive, true negative, false positive, and false negative, respectively \cite{peng2023hierarchical}. Moreover, $ASD = \frac{1}{N}\sum_{i=1}^{N}d(s_{i},g)$, where $N$ is the number of boundary points in the saliency map, $s_{i}$ is the set of boundary points generated by the saliency map, $g$ is the set of ground truth boundary points, and $d(s_{i},g)$ is the distance between the $i$-th boundary point generated by the saliency map and nearest ground truth boundary point. Proportion = $ \frac{\sum_{(i,j) \in bbox} L^{dis}_{(i,j)}}{\sum_{(i,j) \in bbox} L^{dis}_{(i,j)} + \sum_{(i,j) \notin bbox} L^{dis}_{(i,j)}}$, where ``bbox'' denotes the ground truth foreground. $L^{dis}_{(i, j)}$ represents the saliency value of the corresponding saliency map falling on coordinates $(i, j)$.

\subsection{Experimental Details}

All the images in the PALM and EyePACS datasets were cropped to $224 \times 224 \times 3$. ResNet50 \cite{he2016deep} was trained as a weakly supervised NNC. We selected 2,500 images marked as 0 from the EyePACS training set and mix them with the images marked as normal in the PALM training set to form a new normal fundus image training set. We selected 2,500 images labeled as non-0 from the EyePACS dataset and mixed them with images from the PALM training set that were labeled as diseased to create a new diseased fundus image training set. The above normal and diseased fundus image training sets were used as the training set for ResNet50. All parameters of the trained ResNet50 were fixed, and saliency maps were generated on ResNet50 using all baseline methods and HSPI for disease localization testing. We randomly selected 160 images in the PALM test set to participate in the disease localization test. Hyperparameters were set as: $H=W=224$, $\zeta=0.5$, $I=10$, $J=15$, $n^{t} = 10^{4}$, $\eta=10^{-2}$, $H_{i}=W_{i}=6+i$, $i_{b}=1$, $\epsilon = 8$, $\alpha_{(i,j)} = 5 \times 10^{-3}$, $\beta_{(i,j)} = 0.08$, $\lambda_{(i,j)}=1$, $i\in\{1,2,...,I\}$, and $j\in\{1,2,..., J\}$. For quantitative evaluation, we converted the saliency map into a pseudo-mask: positions above the threshold to 1, and positions below the threshold to 0. Our code was implemented using PyTorch, and experiments were conducted on a 2080Ti GPU.

\begin{figure*}[!t]
	\centering
	{\includegraphics[width=1.0\linewidth]{{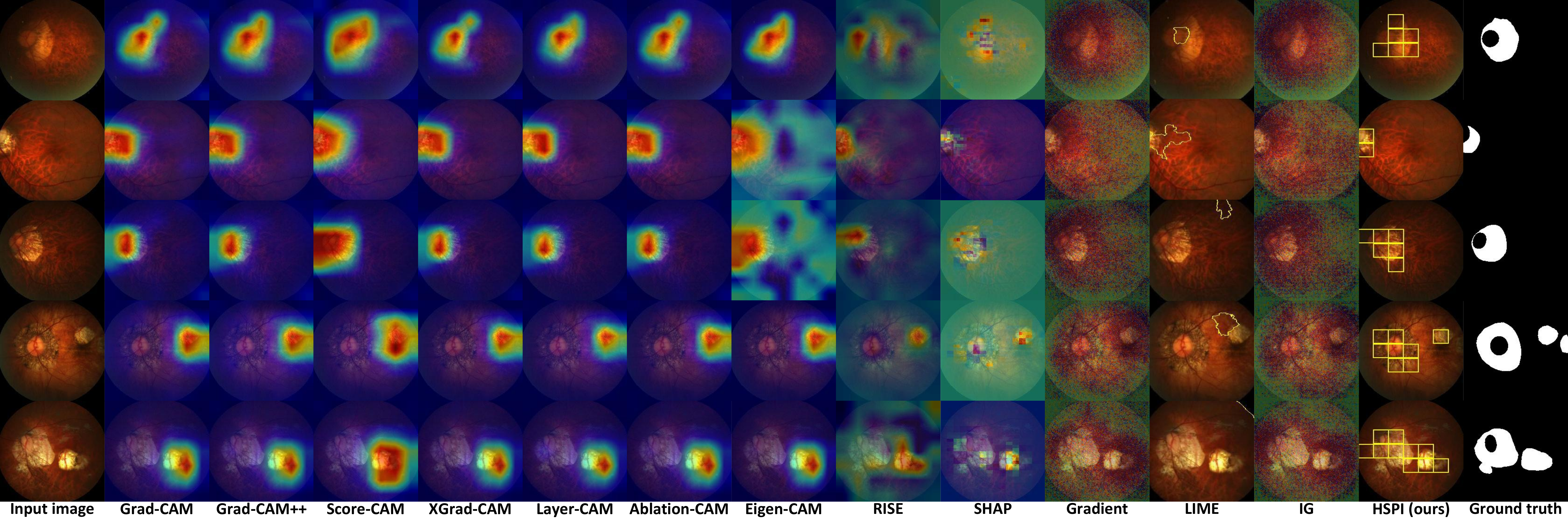}}%
	}
	\caption{Localization results for ResNet50 using different attribution methods. The saliency maps are visualized using JET colormaps.}
	\label{fig_cam}
\end{figure*}

\subsection{Hierarchical Salient Patch Identification Process}

Figure \ref{process} shows how HSPI locates diseases in the input image. In Figure \ref{process}, the first column shows the input image, the ground truth, and the result generated by HSPI. Due to space limitations, only the identification process of the first three prediction-critical patches in the hierarchical identification method is shown. The results on the right show the training vectors, training masks, and heatmaps at different epochs in the HSPI's hierarchical identification. The $7 \times 7$ training vector was upsampled by bilinear interpolation to obtain the training mask. The training mask was mixed with the input image to obtain the heatmap. The training period Epoch, similarity loss $L_{s}$, mask loss $L_{m}$, and maximum-second difference $D_{n}$ corresponding to these images are marked below each group of three images. The gray grid lines in all images are auxiliary line segments to facilitate observation and are not related to the results generated by HSPI. The yellow patch in the candidate image indicates the most critical patch for prediction identified in the input image by SPI of the current stage.

In Stage 1, SPI trained the mask vector by continuously optimizing the consistency loss. When Epoch $= 0$, the training vector was not trained; therefore, the value of each element was 1, all squares were white, and the corresponding training mask and heatmap did not indicate the areas contributing most to predictions. As the training epochs continued to increase, the training vector continued to remove areas irrelevant to prediction by changing the element value to 0 and turning the square black. At Epoch $= 3600$, because $L_{s}$, $L_{m}$, and $D_{n}$ satisfied CPF, the position with the largest value in the training vector was selected as the candidate area (yellow patch).

In Stage 2, the input image and training vector completely masked the candidate area identified in Stage 1 by setting this area to 0. The above process was repeated to train the vector until the CPF was satisfied, and the position of the largest value in the training vector was selected as the candidate area. The processing in Stage 3 was the same. In Figure \ref{process}, the SPI at each stage accurately located the diseased patch, and the HSPI hierarchically utilized SPI to comprehensively identify the diseased patches. Therefore, HSPI can accurately and comprehensively locate the diseased areas. The SPI at each stage identifies the most important patch for NNC prediction by training mask vector to remove regions irrelevant for prediction, so this process is interpretable.

In summary, Figure \ref{process} comprehensively demonstrates the positioning process of HSPI, imitating the method used by doctors to diagnose. HSPI possesses interpretability and can also localize fundus diseases with accuracy and completeness.

\subsection{Qualitative Evaluation}

This experiment aims to qualitatively compare the fundus disease localization capabilities of saliency maps generated using various attribution methods. We compared twelve cutting-edge or classic saliency map techniques. In Figure \ref{fig_cam}, the images in the first column are the input images; the images in the last column are the ground truths, where white and black areas are diseased and normal areas, respectively; and the other columns show the saliency maps generated by other methods.

In Figure \ref{fig_cam}, compared to CAM-based methods, HSPI offered more comprehensive fundus disease localization. Additionally, compared to other attribution methods, HSPI's fundus disease localization was more accurate and easier to understand. First, SPI constrains the optimization of the consistency loss through CPF, allowing the mask vector to precisely learn the disease area. This makes the positioning of the patch identified by SPI more precise at each stage. Second, by using SPI hierarchically, it is mandatory to identify whether different areas in the image contain diseases so that HSPI could comprehensively locate all disease areas. HSPI combines the positioning accuracy of SPI with CPF and the comprehensiveness of hierarchical identification. It uses PSMI to filter out additionally identified non-disease areas, allowing HSPI to achieve the best interpretable disease positioning.

\begin{table}
	\centering
	\caption{Comparative results of segmentation metrics between HSPI and CAM-based methods on the PALM dataset.}
	\begin{tabular}{lccccc}
		\hline
		Evaluation & F1 & PPV & SP & ASD & Proportion \\
		\hline			
		Grad-CAM \cite{selvaraju2017grad} & 0.41 & 0.39 & 0.85 & 13.2 & 0.32 \\
		Grad-CAM++ \cite{chattopadhay2018grad} & 0.44 & 0.41 & 0.86 & 11.3 & 0.31 \\
		Score-CAM \cite{wang2020score} & 0.43 & 0.41 & 0.85 & 11.9 & 0.32 \\
		Ablation-CAM \cite{ramaswamy2020ablation} & 0.42 & 0.40 & 0.85 & 12.5 & 0.32 \\
		Layer-CAM \cite{jiang2021layercam} & 0.45 & 0.42 & 0.86 & 10.8 & 0.35 \\
		XGrad-CAM \cite{fu2020axiom} & 0.43 & 0.41 & 0.86 & 12.1 & 0.34 \\
		Eigen-CAM \cite{muhammad2020eigen} & 0.40 & 0.38 & 0.85 & 15.3 & 0.33 \\
		Eigen Grad-CAM \cite{duguaesescu2022evaluation} & 0.45 & 0.43 & 0.86 & 10.8 & 0.41 \\	
		HSPI (ours) & \textbf{0.46} & \textbf{0.52} & \textbf{0.93} & \textbf{8.5} & \textbf{0.52} \\
		\hline
	\end{tabular}
	\label{Table1}
\end{table}

\begin{table}
	\centering
	\caption{Comparative results of segmentation metrics between HSPI and other attribution methods on the PALM dataset.}
	\begin{tabular}{lccccc}
		\hline
		Evaluation & F1 & PPV & SP & ASD & Proportion \\
		\hline					
		RISE \cite{petsiuk2018rise} & 0.32 & 0.30 & 0.82 & 18.2 & 0.19 \\
		LIME \cite{ribeiro2016should} & 0.27 & 0.18 & 0.01 & 21.6 & 0.18 \\	
		G-LIME \cite{li2023g} & 0.29 & 0.20 & 0.01 & 19.8 & 0.20 \\
		Gradient \cite{simonyan2013deep} & 0.24 & 0.24 & 0.81 & 15.1 & 0.17 \\
		IG \cite{sundararajan2017axiomatic} & 0.23 & 0.23 & 0.81 & 14.9 & 0.17 \\		
		DeepLIFT \cite{shrikumar2017learning} & 0.24 & 0.24 & 0.82 & 14.8 & 0.18 \\	
		SHAP \cite{lundberg2017unified} & 0.31 & 0.28 & 0.66 & 16.4 & 0.18 \\
		HSPI (ours) & \textbf{0.46} & \textbf{0.52} & \textbf{0.93} & \textbf{8.5} & \textbf{0.52} \\
		\hline
	\end{tabular}
	\label{Table2}
\end{table}

\subsection{Quantitative Evaluation}

Tables \ref{Table1} and \ref{Table2} compare the performance of the pseudo-masks generated by the proposed HSPI method with mainstream CAM-based methods and other attribution methods for fundus disease localization. These results showed that the pseudo-masks generated by HSPI performed well on the task of localizing fundus diseases. The proposed HSPI method had significant improvements over previous methods for the PPV, ASD, and Proportion indicators. Because fundus diseases are often irregular or scattered, previous attribution methods can identify only one disease area in an image. In contrast, HSPI can identify disease areas more comprehensively and accurately, effectively avoiding the problems of missed and false detections.

\subsection{Ablation Study}

\begin{table}
	\centering
	\caption{Ablation experiment results used to verify the effectiveness of CPF, hierarchical identification, and PSMI.}
	\begin{tabular}{lccccc}
		\hline
		Method & Uses CPF & Uses PSMI & F1 & PPV & ASD  \\
		\hline
		SPI & & & 0.12 & 0.45 & 28.1  \\		
		SPI & $\checkmark$ & & 0.14 & 0.49 & 23.9  \\
		HSPI (J=2) & $\checkmark$ & & 0.22 & 0.48 & 18.5 \\
		HSPI (J=3) & $\checkmark$ & & 0.27 & 0.47 & 15.3 \\
		HSPI (J=4) & $\checkmark$ & & 0.31 & 0.46 & 13.7 \\
		HSPI (J=5) & $\checkmark$ & & 0.34 & 0.45 & 13.1 \\
		HSPI (J=10) & $\checkmark$ & & 0.42 & 0.41 & 13.2 \\
		HSPI (J=15) & $\checkmark$ & & 0.42 & 0.35 & 14.8 \\
		HSPI (J=15) & $\checkmark$ & $\checkmark$ & \textbf{0.46} & \textbf{0.52} & \textbf{8.5} \\
		\hline
	\end{tabular}
	\label{Ablation}
\end{table}

\subsubsection{Effectiveness of CPF}

Table \ref{Ablation} shows the effectiveness of using CPF in SPI through a quantitative comparison. The first row of data shows the results of SPI without CPF and PSMI, and the second row shows the results of SPI using CPF but not PSMI. After the introduction of CPF, SPI showed improvements in all three indicators and could more accurately locate diseased areas. Figure \ref{process} illustrates the effects of CPF. In Stages 1, 2, and 3, only when Epoch $ = 3600$ do $L_{s}$, $L_{m}$, and $D_{n}$ satisfy CPF. At this point, the training vector accurately locates the disease patch. When CPF is not met, the training vector's localization is relatively inaccurate.

In summary, optimizing $L_{s}$ such that its value is less than a small threshold ensures that the NNC has a consistent output response to the training and input images, so that the training image retains the prediction-critical areas in the input image. Optimizing $L_{m}$ to be less than a small threshold ensures that the training vector removes parts that are less relevant to the prediction. Selecting the maximum $D_{n}$ ensures that the training vector forms a prominent peak so that the patch most crucial for NNC prediction can be visually identified. By applying these three constraints of CPF, SPI can locate the fundus disease area more accurately.

\subsubsection{Effectiveness of Hierarchical Identification}

Table \ref{Ablation} illustrates the role of hierarchical identification. The second row indicates that hierarchical identification was not used, while the third to eighth rows indicate that hierarchical identification was used. $J$ is the number of hierarchical levels. Both SPI and HSPI used CPF but not PSMI. As shown in Table \ref{Ablation}, as $J$ increases, F1 gradually increases, indicating that the disease localization becomes increasingly comprehensive. Because SPI identifies prediction-critical areas at the patch level, using SPI once may not fully cover all diseased areas. HSPI improves this situation using hierarchical identification. In Figure \ref{process}, SPI was used to analyze different areas in the first, second, and third stages. This ensured that the identified patches covered all disease areas and avoided missing disease areas caused by a single application of SPI. This highlights the effectiveness of hierarchical identification.

\subsubsection{Effectiveness of PSMI}

The penultimate and final rows in Table \ref{Ablation} show the positioning effect of HSPI ($J$ = 15) without and with PSMI, respectively. Table \ref{Ablation} demonstrates that after using PSMI, HSPI improved across all indicators, resulting in more accurate and comprehensive positioning. Thus, PSMI can effectively fuse the identification results of mask vectors of different sizes, thereby eliminating misidentification or redundant patches. This allows HSPI to more accurately locate fundus disease areas relevant to prediction.

\section{Conclusion}
In this study, we proposed and validated a weakly supervised, interpretable fundus disease localization method called HSPI. This method uses image-level labels and an NNC to achieve interpretable, comprehensive, and accurate localization of fundus diseases. Through innovative SPI technology, we can identify and analyze an image patch that is critical to NNC prediction. In addition, the hierarchical identification strategy we introduced enables HSPI to identify different patches to ensure a more comprehensive coverage of potential disease areas. Our proposed CPF and PSMI further improved the accuracy of positioning and excluded additional identification of non-diseased areas, effectively improving the diagnostic performance of the HSPI. Experimental results on fundus image datasets show that the positioning capabilities of HSPI are better than those of existing explainable attribution methods for multiple evaluation indicators. Additional ablation studies confirmed the effectiveness of the proposed methods. In summary, the proposed HSPI is a novel and effective tool for the interpretable diagnosis and research of fundus diseases.

\end{document}